\documentclass[letterpaper, 10 pt, conference]{ieeeconf}  

\IEEEoverridecommandlockouts                              

\usepackage{graphics} 
\usepackage{epsfig} 
\usepackage{mathptmx} 
\usepackage{times} 
\usepackage{amsmath} 
\usepackage{amssymb}  
\usepackage{algorithm}
\usepackage{algpseudocode}
\usepackage{booktabs}
\usepackage{multirow}
\usepackage{graphicx}
\usepackage{relsize}
\usepackage{subcaption}
\usepackage{mathtools, nccmath}
\usepackage[section]{placeins}
\usepackage{wrapfig}
\usepackage{xcolor}
\usepackage{diagbox}
\usepackage{ulem}
\usepackage{hyperref}
\usepackage{cite}
\usepackage{url}

\title{\LARGE \bf
How Vulnerable Is My Learned Policy? \\Adversarial Attacks On Modern Behavior Cloning Algorithms
}

\author{Akansha Kalra$^{*,1}$, Basavasagar Patil$^{*,2}$, Guanhong Tao$^{1}$, and Daniel S. Brown$^{1}$
\thanks{*Equal Contribution}
\thanks{$^{1}$University of Utah,
        {\tt\small akansha.kalra@utah.edu}}%
\thanks{$^{2}$University of California, Irvine
        }%
}

\begin{document}

\maketitle
\thispagestyle{empty}
\pagestyle{empty}

\begin{abstract}
Imitation learning, also known as learning from demonstrations, is a popular approach to train AI models; however, the vulnerability of these models to adversarial attacks remains underexplored. 
We present the first systematic study of adversarial attacks, across a range of both classic and recently proposed imitation learning algorithms, including Vanilla Behavior Cloning (Vanilla BC), LSTM-GMM, Implicit Behavior Cloning (IBC), Diffusion Policy (DP), and Vector-Quantized Behavior Transformer (VQ-BET).
We study the vulnerability of these methods to white-box, grey-box and black-box adversarial perturbations. 
Our experiments reveal that most existing methods are highly vulnerable to these attacks, including black-box transfer attacks that transfer across algorithms. 
White-box attacks cause at least a 65\% reduction in average task success across all evaluated tasks and algorithms, while the black-box transfer attacks reduce task success by up to 88\% on Lift, 99\% on Can, and 100\% on Square.
To the best of our knowledge, we are the first to study and compare the vulnerabilities of different popular imitation learning algorithms to both white-box and black-box attacks. Our findings highlight the vulnerabilities of modern imitation learning algorithms, paving the way for future work in addressing such limitations. Videos and code are available at \url{https://sites.google.com/view/uap-attacks-on-bc}.
\end{abstract}


\section{INTRODUCTION}

Imitation learning has emerged as a powerful paradigm in AI and robotics, enabling agents to learn complex behaviors from expert demonstrations~\cite{zare2024survey}. 
Imitation learning via supervised learning approaches is known as behavior cloning~\cite{torabi2018behavioral}.
Behavior cloning (BC), sometimes also called supervised fine-tuning, is increasingly being deployed in real-world scenarios, such as robots in industrial automation~\cite{kernbach2026behavioral} and household robotics~\cite{shafiullah2023bringing} and large language models~\cite{ouyang2022training}.
However, the degree to which these policies pose potential security risks and their vulnerability to manipulation by adversaries to cause undesired behaviors or even catastrophic incidents is largely unknown. 
Motivated by these potential risks, we perform the first study of the vulnerabilities of a variety of different BC algorithms to adversarial attacks.

Adversarial attacks are a widely studied area in machine learning that aim to develop imperceptible perturbations to change the output of machine learning models~\cite{Szegedy2013IntriguingPO,Goodfellow2014ExplainingAH,chakraborty2021survey}.
However, the robustness of BC models to adversarial attacks has been largely overlooked in prior research, particularly in the context of robotic manipulation tasks. Prior works either only analyze the vulnerability of a single type of BC algorithm in isolation while only considering white-box attacks~\cite{hall2020studying,wu2023adversarial,boloor2019simple,Chen2024DiffusionPA} or investigate attacks in RL settings ~\cite{Mo2023AttackingDR, Sun2020StealthyAE,Pattanaik2017RobustDR,he2023quantifying}, where an AI agent learns from expected reward in the environment rather than learning purely from human demonstrations in absence of a reward signal.
By contrast, we perform the first investigation of the relative vulnerabilities of different BC algorithms under both white- and black-box attacks via universal adversarial perturbations. 
In our work we first examine post-deployment white-box attacks, where an adversary has access to the trained model parameters but cannot modify the training process.
Unlike dataset attacks that aim to corrupt the learning process~\cite{goldblum2022dataset,kalra2025dataset}, our attacks target the inference phase, attempting to cause task failures through perturbations to visual observations. 
While prior work focuses on attacks that require expensive test-time optimization of pixel perturbations per input state~\cite{hall2020studying,wu2023adversarial,boloor2019simple,Chen2024DiffusionPA}, we demonstrate the effectiveness of universal adversarial attacks~\cite{MoosaviDezfooli2016UniversalAP}---attacks that optimize a single perturbation that is designed to work across the entire input distribution---across a variety of BC algorithms. These attacks are efficient to implement since they do not require any test-time optimization and shed light on the susceptibility of popular BC algorithms to adversarial attacks.
In particular, we evaluate the adversarial robustness of five popular BC frameworks: Vanilla Behavior Cloning (Vanilla BC)~\cite{pomerleau1988alvinn}, Long Short-Term Memory with Gaussian Mixture Model (LSTM-GMM) ~\cite{Mandlekar2021WhatMI}, Implicit Behavior Cloning (IBC)~\cite{florence2021implicit}, Diffusion Policy (DP)~\cite{chi2023diffusionpolicy}, and Vector Quantized Behavior Transformer (VQ-BET)~\cite{Lee2024BehaviorGW} and find that they are all highly susceptible to offline-generated, universal adversarial attacks.

Additionally, while prior work~\cite{hall2020studying,wu2023adversarial,boloor2019simple,Chen2024DiffusionPA} has only considered white-box attacks, we perform the first study of adversarial black-box transfer attacks across different BC algorithms.
Overall, we hope our results serve as an impetus for enhancing awareness of the security and reliability concerns regarding policies learned via behavior cloning as these approaches gain more widespread use, especially in cyber-physical systems.

\begin{figure*}
    \centering
      \includegraphics[width=0.9\linewidth]{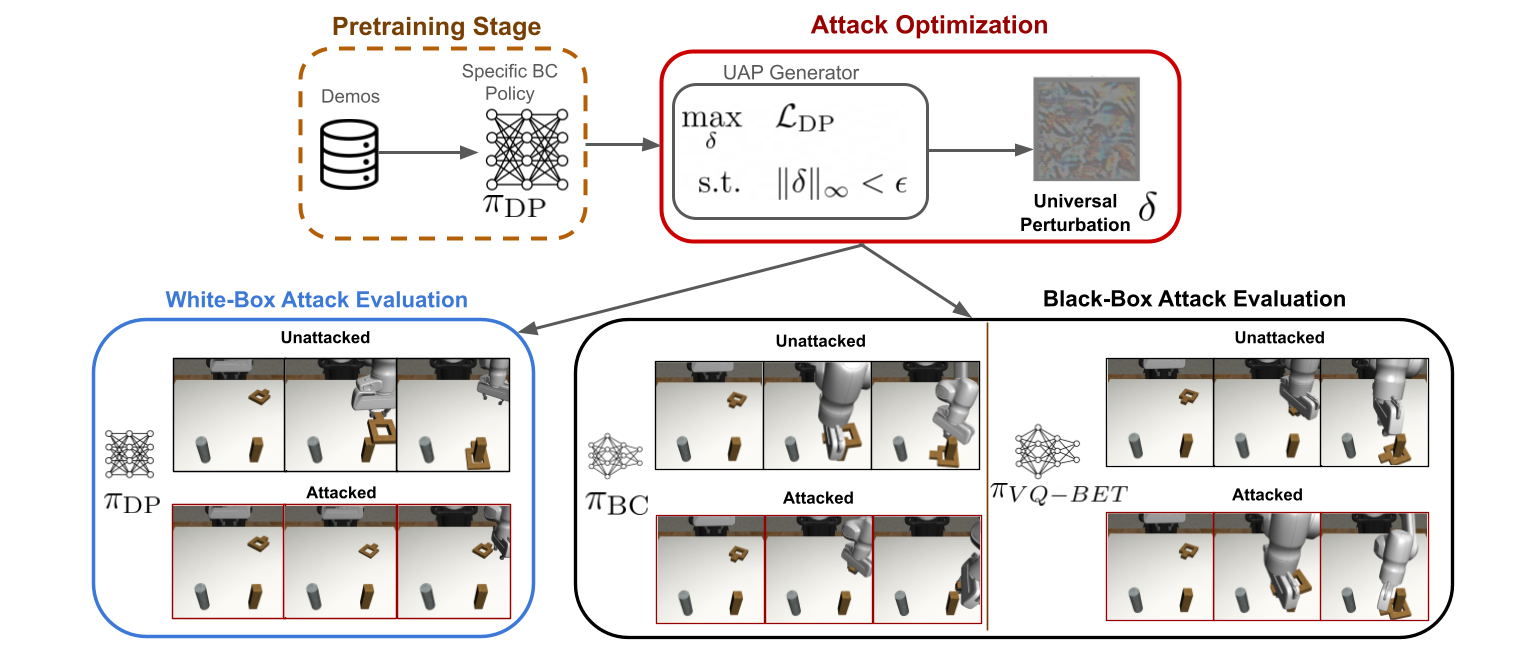}
    \label{fig:pipeline}
    \captionsetup{width=0.95\textwidth}
    \caption{\textbf{Universal Adversarial Perturbation (UAP) Attack Pipeline for Behavior Cloning (BC) Algorithms.} 
We craft adversarial attacks on pre-trained behavior cloning (BC) policies (illustrated here using Diffusion Policy $\pi_{DP}$). 
By applying learned universal adversarial attacks at test time via perturbed visual observations, we test the robustness of BC policies to both white-box and black-box attacks. 
In the white-box setting (bottom left), the attack degrades the performance of the source policy $\pi_{DP}$. 
In the black-box setting (bottom right), 
an attack crafted for $\pi_{DP}$ is applied to different target policies (e.g., $\pi_{BC}$ and $\pi_{VQ\text{-}BET}$). 
Unattacked and attacked rollouts shown here illustrate successful black-box transfer  attack to $(\pi_{BC}$), causing task failure, but limited transfer to $(\pi_{VQ\text{-}BET})$.}

    %
    
    \vspace{-3mm}
\end{figure*}

The primary contributions of our work are as follows:
\begin{itemize}
    \item We conduct the first comparison of white-box adversarial attacks across five popular and widely used behavior cloning algorithms, Vanilla BC, LSTM-GMM, IBC, DP, and VQ-BET. We find that even the most modern state-of-the-art algorithms are vulnerable to attack.

    \item We provide the first study of black-box transfer attacks on BC policies and provide insights into the transferability of attacks across algorithms.
    We find surprising transferability of the attacks, highlighting the importance of considering robustness in the training pipeline of modern imitation learning.
    

    \item We present a systematic evaluation of the robustness of BC policies to universal adversarial perturbations across multiple threat models: white-box attacks to measure worst-case vulnerability under complete access to a pre-trained BC model, grey-box transfer to quantify within-algorithm vulnerability, and black-box transfer attacks that measure cross-algorithm vulnerability. 
    We find that implicit algorithms such as IBC and Diffusion Policy, and transformer-based policies such as VQ-BET, demonstrate comparatively greater resilience across all settings, though all methods remain highly vulnerable and suffer significant degradation in performance, particularly under larger perturbation budgets and more complex tasks.

\end{itemize}

\section{RELATED WORK}
Adversarial attacks have been well-studied in Machine Learning (ML) with the objective of developing imperceptible perturbations to change the output of the ML models.
Early work by Szegedy et al. \cite{Szegedy2013IntriguingPO} and Goodfellow et al. \cite{Goodfellow2014ExplainingAH} revealed that adding small, imperceptible perturbations to images could drastically alter the prediction of neural network classifiers.
Since then, a significant number of works have explored various adversarial attack methods~\cite{akhtar2018threat,zhang2020adversarial,chakraborty2021survey}.

However, there has been relatively little prior work exploring adversarial attacks on behavior cloning algorithms:
Hall et al.~\cite{hall2020studying} and Boloor et al.~\cite{boloor2019simple} demonstrate that Vanilla BC policies are vulnerable to adversarial perturbations in driving simulations while Wu et al.~\cite{wu2023adversarial} explore attacks against a Vanilla BC policy trained on image observations, and the most recent work by
Chen et al.~\cite{Chen2024DiffusionPA}, explores adversarial attacks on diffusion policies~\cite{chi2023diffusionpolicy}. 
However, all of these works are limited to only online white-box attacks and solely focus on a single type of BC  algorithm.
By contrast, we consider universal attacks in both white-box and black-box settings across multiple BC algorithms and present the first systematic study of adversarial robustness and cross-algorithm transferability in behavior cloning policies. By studying how well adversarial perturbations transfer across different BC algorithms, our results offer a broader and more comprehensive perspective on the vulnerability of modern behavior cloning approaches.


There is a large body of related work that focuses on adversarial attacks on reinforcement learning (RL) algorithms~\cite{Gleave2019AdversarialPA,zhang2021robust,moos2022robust,casper2022white,chen2023bird,Mo2023AttackingDR}; however, these attacks are specifically designed for RL settings where policies are trained to maximize expected reward, and adversarial perturbations are evaluated or optimized based on their impact on the expected discounted sum of rewards. In contrast, our work is situated entirely in the imitation learning (IL) setting, where no reward signal is available, and policies are learned purely from demonstrations with no access to a simulator for trial and error learning. As such, the core mechanisms in RL attack methods---including adversarial value estimation, robust policy gradients, and repeated environment interactions for adversarial training---do not apply in our setting.


\section{Preliminaries}

\subsection{Background and Notation}

Behavior Cloning (BC) aims to learn a policy from expert demonstrations in an environment modeled as a Markov Decision Process (MDP), with state space $S$, action space $A$, transition function $T:S\times A \rightarrow S$, and reward function $R$. However, unlike in standard planning or reinforcement learning problems~\cite{sutton1998reinforcement}, we focus on the imitation learning setting~\cite{osa2018algorithmic} where both the transition function and reward function are unknown and the policy must be learned by observing an expert act in the environment.
We denote a dataset of expert demonstrations as $\mathcal{D} = \{ \tau_1, \tau_2, \dots, \tau_N \}$, where each trajectory $\tau = (s_0, a_0, s_1, a_1, \dots)$ consists of a sequence of states and actions generated by the expert. 
Behavior cloning is a supervised learning framework that learns a policy $\pi_{\theta}(s)$, parameterized by $\theta$, to imitate expert behavior. We discuss the specifics of the BC algorithms we study in Section~\ref{sec:methodology}.

\subsection{Universal Adversarial Perturbation Attacks} 
Moosavi-Dezfooli et al. ~\cite{MoosaviDezfooli2016UniversalAP} proposed the Universal Adversarial Perturbation (UAP) attack, a popular class of adversarial attacks that learn a single, input-agnostic, vector perturbation $\delta$ to cause misclassifications across multiple different inputs.
Unlike online adversarial attacks, such as Fast Gradient Sign Method (FGSM)~\cite{Goodfellow2014ExplainingAH} or Projected Gradient Descent (PGD)~\cite{madry2018towards} attacks, which compute perturbations for each input independently and require expensive white-box access at test-time, UAP generates a single perturbation $\delta$ that satisfies $\|\delta\|_p \leq \epsilon$ and is designed offline to successfully affect a high fraction of the input distribution. 
The perturbation is computed iteratively over a set of training inputs by accumulating perturbations that increase the loss of the pre-trained neural network for each input while maintaining the constraint on the perturbation magnitude.

\section{Problem Settings}

We focus on the following three settings and corresponding research questions: 
\begin{enumerate}
    \item \textbf{White-Box Attacks:} In this setting, the attacker has white-box access to the trained policy parameters but cannot modify them. Given access to a trained BC algorithm, we study whether an attacker can learn a minimal universal adversarial perturbation that, when applied at test time, leads to task failure.

     \item \textbf{Grey-Box Transfer Attacks:} Under this setting, the attacker has knowledge about the type of BC algorithm and the neural network architecture but has no access to the exact model parameters.
     The adversary constructs the UAP attack using a surrogate model of the known BC algorithm type and the model architecture.
     We evaluate transferability of adversarial perturbations across independently trained policies of the same algorithm to quantify within-algorithm vulnerability.
     
    \item \textbf{Black-Box Transfer Attacks:} In this setting, the attacker requires no access to the target model's parameters or training data, making these attacks feasible in black-box scenarios. 
    We perform black-box transfer attacks by applying an attack generated for a source BC model with known weights to a different unknown target BC model with unknown weights. We study whether black-box transfer attacks are successful across different source and target BC algorithms.
    
\end{enumerate}

\section{UAP attacks on BC algorithms}\label{sec:methodology}

We consider a threat model which is motivated by the increasing deployment of pre-trained AI models with publicly released parameters and with no downstream retraining required.
In particular, we analyze offline Universal Adversarial Perturbation (UAP) attacks on these pre-trained BC models. 
The goal of these attacks is to find a single fixed perturbation which generalizes across all input states, such that when applied during inference to these pre-trained models, it induces task failure. 

We formulate the universal perturbation search as the following maximization problem: 
\begin{equation}
\max_{\|\delta\|_\infty \leq \varepsilon} 
\ \mathbb{E}_{(s,a)\sim\mathcal{D}}
\!\Big[
\mathcal{L}\big(\pi_\theta(s+\delta), a \big)
\Big],
\end{equation}
where $\pi_\theta$ denotes the pre-trained behavior cloning policy, $s$ refers to the image observation, $a$ is the ground-truth  action, and $\mathcal{L}$ represents the loss computed according to the choice of underlying behavior cloning.
This approach is particularly relevant for real-world scenarios where computing per-input perturbations may not be feasible, and the same attack vector needs to remain effective across a wide range of observations. 
\begin{algorithm}[!t]
\caption{Universal Adversarial Perturbation (UAP) Attack for BC Algorithms}
\label{alg:uap}
\begin{algorithmic}[1]
\Require 
 Dataset $\mathcal{D} = \{ \tau_1, \tau_2, \dots, \tau_N \}$, where trajectory $\tau = (s_0, a_0, \dots)$
 sequence of states $s$ and actions $a$, 
pre-trained BC policy $\pi_\theta$, perturbation budget $\varepsilon$, step size $\alpha$, epochs $E$ 


\State Initialize universal perturbation $\delta = 0$
\For{epoch $e = 1$ to $E$}
  \State Reset gradient accumulator $G = 0$
  \For{each batch $(s, a)$ from $\mathcal{D}$}
    \State Generate perturbed input state $\tilde{s} = \mathrm{clip}(s + \delta, 0, 1)$
    \State Predict action $\hat{a} = \pi_\theta(\tilde{s})$
    \State Compute $\mathcal{L}_{BC}(\tilde{s},\hat{a})$ according to BC policy
    \State Accumulate input sensitivity: $G \mathrel{+}= \nabla_{\tilde{s}} \mathcal{L}_{BC}$
  \EndFor
  \State Update perturbation: $ \delta \gets \mathrm{clip}_{[-\varepsilon,\varepsilon]}\!\big(\delta + \alpha \cdot \mathrm{sign}(G)\big)$
\EndFor
\State \Return $\delta$
\end{algorithmic}
\end{algorithm}

Algorithm \ref{alg:uap} presents a unified UAP attack framework  that applies to all BC models considered in this paper.
Given a dataset $\mathcal{D}$ of state–action pairs, a perturbation budget $\epsilon$, step size $\alpha$, and epochs $E$, we optimize a single universal perturbation $\delta$ that maximizes the policy-specific training loss. At each epoch, perturbed states $\tilde{s}=\mathrm{clip}(s+\delta,0,1)$ are passed through $\pi_\theta$, gradients of the relevant loss function are accumulated, and $\delta$ is updated under the $L_\infty$ constraint $|\delta|_\infty\leq \epsilon$.

We evaluate adversarial attacks across five popular BC algorithms. 
We first briefly define each algorithm and then show how Algorithm 1 is instantiated to generate a universal adversarial perturbation attack based on the respective algorithm. 
Perturbations are limited to the visual observation space and no direct action manipulation is allowed. 
Perturbations must remain within an $L_\infty$ norm ball of radius $\epsilon$ to maintain imperceptibility. 
The attacker can compute gradients through the entire policy network. 
While the input to the BC policy includes both low-dimensional proprioceptive signals and high-dimensional visual observations, we perturb only the vision component of the hybrid state following prior works ~\cite{Chen2024DiffusionPA,hall2020studying,boloor2019simple}, which demonstrate the effectiveness of adversarial perturbations on visual inputs in behavior cloning settings. 



\paragraph{\textbf{Vanilla Behavior Cloning (Vanilla BC)}} 
Vanilla BC learns a policy via standard supervised learning~\cite{bain1995framework,torabi2018behavioral}. Given a dataset of expert demonstrations it directly maps states to actions using a neural network, $\pi_\theta$ trained to minimize the mean squared error (MSE) for continuous actions:
\begin{equation}
\label{eq:bc_loss}
\mathcal{L}_{\text{BC}}(\theta)
= \mathbb{E}_{(s,a)\sim\mathcal{D}} \| \pi_\theta(s) - a   \|_2^2
\end{equation}

For Vanilla BC, the adversarial attacks aim to maximize the mean squared error (MSE) loss (Equation~\eqref{eq:bc_loss}) between the predicted and expert actions by introducing small perturbations to the input states.

\paragraph{\textbf{Long Short-Term Memory with Gaussian Mixture Model (LSTM-GMM)}} The LSTM-GMM algorithm augments Vanilla BC with a Recurrent Neural Network, specifically a Long Short-Term Memory (LSTM)~\cite{Hochreiter1997LongSM}, to encode temporal dependencies and a Gaussian Mixture Model (GMM) ~\cite{mclachlan2000finite} to model the multimodal action distribution~\cite{Mandlekar2021WhatMI}.
Given a demonstrated sequence of states $(s_1,\ldots,s_T)$, the LSTM network maintains a hidden state $h_t$ to capture past state observations, that is then used as input to the policy $\pi_\theta(a_t|s_t,h_{t-1})$ to predict a multimodal action distribution: $p(a_t|s_t, h_{t-1}, \theta) = \text{GMM}(h_t)$. 
The policy is trained by maximizing the likelihood of the observed actions given the state sequence: \begin{equation}\label{eq:lstm-gmm-loss}
    \mathcal{L}_{\rm BC\_RNN}(\theta) = \mathbb{E}_{\tau \sim \mathcal{D}} \sum_{t=1}^T -\log p(a_t | s_t, h_{t-1}, \theta).
\end{equation}

For LSTM-GMM, our UAP attack targets the temporal dependencies modeled by the LSTM and the multimodal action distributions captured by the Gaussian Mixture Model (GMM), with the goal of degrading the likelihood maximization assigned to expert actions by the GMM outputs (Equation~\eqref{eq:lstm-gmm-loss}).


\paragraph{\textbf{Implicit Behavior Cloning (IBC)}} IBC reformulates policy learning as energy-based modeling~\cite{florence2021implicit}.
Instead of directly predicting actions, IBC learns an energy function $E_\theta(s, a)$, that assigns low energy to expert actions and high energy to other actions, thereby implicitly defining the policy $\pi(a|s)$. The energy function \(E_\theta({s}, {a})\), parameterized by \(\theta\), is trained using the InfoNCE loss: 
\begin{equation}\label{eq:ibc-loss}
\mathcal{L}_{\text{InfoNCE}} = \mathbb{E}_{(s,a)\sim\mathcal{D}} -\log \left( \frac{e^{-E_\theta\left({s}_i,{a}_i\right)}}{e^{-E_\theta\left({s}_i, {a}_i\right)} + \sum_{j=1}^{N_{\text{neg.}}} e^{-E_\theta\left({s}_i, \tilde{{a}}_i^j\right)}} \right)
\end{equation} 
where $\tilde{{a}}_i^j$ for $j=1,\ldots N_{\text{neg.}}$ are the negative samples {and $i = 1 \ldots N$ are the training samples in the batch.}
This formulation enables IBC to capture complex and multimodal action distributions by searching for the action with minimum energy during inference:
\begin{equation}
\hat{\pi}_{\rm IBC}(s)= \arg\min_{a\in A'} E_{\theta}(s,a)
\end{equation}
where $A'$ is a set of actions sampled from the action space.
Florence et al~\cite{florence2021implicit} described three different techniques to generate the negative samples $\tilde{{a}}_i^j$ and perform inference over the learned energy model $E_\theta(s, a)$ in IBC.
Following prior work~\cite{chi2023diffusionpolicy}, we use the derivative-free optimizer for inference (refer to ~\cite{florence2021implicit} for complete method). 
To attack IBC, the objective is to perturb the state $s$  by finding a perturbation  $\delta$  that increases the energy (reduces the probability) of the demonstrator's action at that state (Equation~\eqref{eq:ibc-loss}).

\paragraph{\textbf{Diffusion Policy (DP)}}  DP learns a policy $\pi_{\theta}$ parameterized by a denoising diffusion probabilistic model~\cite{Ho2020DenoisingDP} $\epsilon_{\theta}$, conditioned on state $s$.
DP does not model an explicit likelihood $p(a|s)$  but instead models the reverse conditional distribution $p(a^{k-1}|a^k,s)$.
During training, Gaussian Noise $\epsilon \sim \mathcal{N}(0,I)$ is added to actions at each diffusion step $k$, to produce the noisy action $a_t^k$ while the diffusion network predicts the noise $\epsilon_{\theta}(s_t,a_t^k,k)$.
Formally, given the actual unmodified, i.e., un-noised action $a^0_t$ at timestep $t$, we learn the parameters $\theta$ by fitting the conditional noise predictor to predict the learned noise
\begin{equation}\label{eq:dp-loss}
    L = \mathbb{E}_{(s_t,a^0_t)\in \mathcal{D}}\left[
\left\|
\epsilon - \epsilon_\theta(s_t,a_t^k, k)
\right\|^2
\right]
\end{equation}
At inference, DP predicts an action sequence spanning $T_p$ future timesteps, but executes actions in horizon $T_a$ before replanning where $T_a \leq T_p$.
To formulate a universal adversarial attack for DP, we use the MSE loss between the predicted ``denoised" actions on the perturbed observation with the random noise added and the actual added noise perturbation (Equation~\eqref{eq:dp-loss}).

\begin{figure}[t]
    \centering
    \begin{subfigure}[b]{0.31\linewidth}
        \centering
        \includegraphics[width=\linewidth]{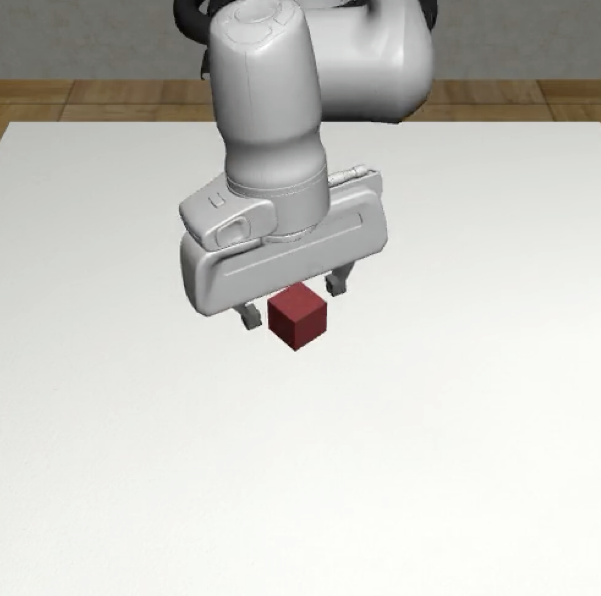}
        \caption{Lift}
        \label{fig:image1}
    \end{subfigure}%
    \hfill
    \begin{subfigure}[b]{0.31\linewidth}
        \centering
        \includegraphics[width=\linewidth]{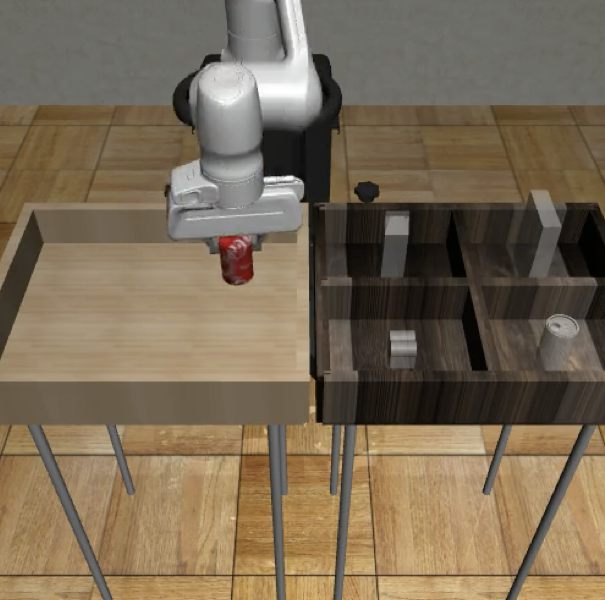}
        \caption{Can}
        \label{fig:image2}
    \end{subfigure}%
    \hfill
    \begin{subfigure}[b]{0.31\linewidth}
        \centering
        \includegraphics[width=\linewidth]{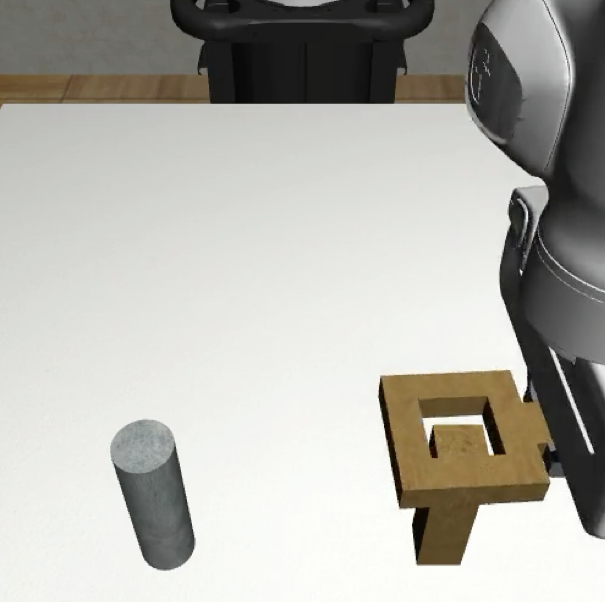}
        \caption{Square}
        \label{fig:image3}
    \end{subfigure}%
    \hfill
    \caption{\textbf{Testing Environments.} We craft and evaluate Universal Adversarial Perturbation attacks to study adversarial robustness of modern behavior cloning algorithms under three RoboMimic~\cite{Mandlekar2021WhatMI} benchmark tasks.}
    \label{fig:robomimic_envs}
\end{figure}

\paragraph{\textbf{Vector Quantized Behavior Transformer (VQ-BET)}} VQ-BET is a latent autoregressive GPT-style transformer model that leverages vector quantization via a hierarchical Variational Autoencoder (VQ-VAE)~\cite{Lee2024BehaviorGW}
for action discretization to capture multimodal behavior.
Instead of directly predicting high-dimensional continuous actions, VQ-BeT first encodes action chunks into a compact latent space, where coarse and fine-grained codes represent multimodal structure. The transformer autoregressively predicts these codes conditioned on past observations (and optional goals), and a lightweight decoder reconstructs actions from the predicted tokens. 

VQ-BET uses a focal loss $\mathcal{L}_{\text {code}}$  to train the code prediction head by comparing the probabilities of the predicted
categorical distribution with the sequence of chosen
codes at each  vector quantized codebook (see \cite{Lee2024BehaviorGW} for the full loss formulation). 
An additional offset head refines the decoded actions for fidelity, by measuring L1 distance between the predicted action and ground truth action. Thus, $\mathcal{L}_{\mathrm{VQ}-\mathrm{BeT}}$ is defined as: 
\begin{equation}\label{eq:vq-bet-loss}
    \mathcal{L}_{\mathrm{VQ}-\mathrm{BeT}}(\theta)=\mathcal{L}_{\text {code }}(\theta)+\left\|a_{t}-\pi_\theta^{resid}(s_t)\right\|_1
\end{equation}
where $a_{t}$ and $\pi_\theta^{resid}(s_t)$ are ground truth and predicted residual action chunks of size $n$, respectively.
For VQ-BET attacks, we exploit the latent action space by targeting the prediction loss of discrete latent codes (Equation~\ref{eq:vq-bet-loss}), ultimately leading to suboptimal action predictions.

\section{EXPERIMENTS \& RESULTS}
We design our experiments to answer the following questions:
    (1) How vulnerable are different behavior cloning algorithms to adversarial attacks?
    (2) How sensitive are attacks to the adversarial perturbation budget?
    (3) How transferable are the attacks across different BC algorithms? 
    
\begin{table*}[t!]
\centering
\caption{\textbf{Robustness of BC policies to UAP attacks}: Task success rates measured as Mean ± Std over 3 seeds with 50 different
environment initial conditions per seed (150 in total) under unattacked and UAP attack for Vanilla BC, LSTM-GMM, IBC, DP, and VQ-BET across all three tasks starting from left to right: Lift, Can, and Square task. Our results demonstrate that classical explicit BC methods (Vanilla BC and LSTM-GMM) are highly vulnerable, while implicit methods (IBC and DP) and recent transformer-based modern explicit BC algorithm (VQ-BET) maintain comparatively higher success on some of the simpler tasks, but are still highly vulnerable, especially for more complex tasks like Can and Square.}
\label{table:UAP_attack_results}
\renewcommand{\arraystretch}{1.2}
\begin{tabular}{|l|cc|cc|cc|}
\hline
\diagbox{{\textbf{BC Policy}}}{{\textbf{Task}}}
& \multicolumn{2}{c|}{\textbf{Lift}} 
& \multicolumn{2}{c|}{\textbf{Can}} 
& \multicolumn{2}{c|}{\textbf{Square}} \\ \hline

& \textbf{Unattacked} & \textbf{Attacked} 
& \textbf{Unattacked} & \textbf{Attacked}
& \textbf{Unattacked} & \textbf{Attacked} \\ \hline
\textbf{Vanilla BC} 
& 0.97(0.02) &0.07(0.08)
& 0.76 (0.18) & 0.00 (0.00)
& 0.50 (0.07) & 0.00 (0.00) \\ \hline

\textbf{LSTM-GMM} 
& 0.95 (0.08) & 0.00 (0.00)
& 0.90 (0.05) & 0.05 (0.06)
& 0.51 (0.09) & 0.00 (0.00) \\ \hline

\textbf{IBC} 
& 0.76 (0.10) & 0.69 (0.10)
& 0.02 (0.02) & 0.01 (0.02)
& 0.00 (0.00) & 0.01 (0.01) \\ \hline

\textbf{DP} 
& 1.00 (0.00) & 0.47 (0.34) 
& 0.97 (0.03) & 0.00 (0.00)
& 0.96 (0.00) & 0.01 (0.02) \\ \hline

\textbf{VQ-BET} 
& 1.00 (0.00) & 0.41 (0.35) 
& 0.97 (0.02) & 0.01 (0.02)
& 0.59 (0.11) & 0.02 (0.02) \\ \hline
\end{tabular}
\end{table*}
\subsection{Environments}

To test the adversarial robustness of modern BC algorithms, we consider a set of common benchmarks shown in Figure~\ref{fig:robomimic_envs}. 
 Lift, Can and Square are tasks taken from Robomimic \cite{Mandlekar2021WhatMI}, where the state-of-the-art frameworks such as Diffusion Policy and LSTM-GMM have been shown to have a nearly 100\% success rate in non-adversarial settings.
All three tasks---Lift, Can and Square---have a 7D action space with 84x84 image observations. 

\textbf{Lift} is a task where a robot arm must lift a small cube from a table surface. Success is determined by elevating the cube above a threshold height. Initial cube poses are randomized with z-axis rotation within a small square region at the table center.

\textbf{{Can}} requires the robot to transfer a soda can from a large source bin into a smaller target bin. 
This task presents increased difficulty over Lift due to the more complex grasping requirements of the cylindrical can and the constrained placement target. 
The can's initial pose is randomized with z-axis rotation anywhere within the source bin.

\textbf{{Square}} is a high-precision manipulation task where the robot must pick up a square nut and insert it onto a vertical rod. This task significantly increases complexity by requiring precise alignment and complex insertion dynamics. The nut's initial pose is randomized with z-axis rotation within a square region on the table surface.
Initial positions of both the square-shaped block and the end-effector are randomized to ensure learned policies must generalize across different pushing strategies.

\subsection{Pre-trained Policies}
To provide consistent and reproducible results, we attack the pre-trained checkpoints for LSTM-GMM, IBC, and DP released by the authors of Diffusion Policy~\cite{chi2023diffusionpolicy} for this suite of tasks, and train our policies for Vanilla-BC and VQ-BET, due to the absence of publicly available checkpoints. To facilitate reproducibility, we release all code and policy checkpoints on our project website\footnote{ \url{https://sites.google.com/view/uap-attacks-on-bc}}.
For each environment, we evaluate performance on 50 randomly initialized instances across three seeds and report the mean and standard deviation of the task success rate.
For our attacks on Lift, Can, and Square we use the proficient human demonstration datasets from Robomimic \cite{Mandlekar2021WhatMI} and attack observations from both camera views: front-view camera and the wrist-mounted camera. Following recent work on adversarial attacks~\cite{wu2024adversarial}, we use an overall attack budget of $\varepsilon = 0.0625$ (16/256) and clamp the universal perturbation generated to be in the range $[-\epsilon,\epsilon]$. 
We also clip the perturbed observations (observations with universal perturbations added) to be in $[0,1]$ range.
Additional details on the experimental setup, evaluation metrics, and implementation are available on the website\footnotemark[\value{footnote}].


\subsection{\textbf{How vulnerable are different behavior cloning algorithms to adversarial attacks?}} 


To establish the worst-case vulnerability baseline under complete access to the pre-trained model, we first evaluate each modern BC algorithm under offline UAP attacks learned in a white-box setting, before studying black-box transferability.
\begin{figure*}[t]
    \centering

    \begin{subfigure}[t]{0.48\textwidth}
        \vspace{0pt}
        \centering
        \parbox[c][5.2cm][c]{\linewidth}{%
            \centering
            \includegraphics[
                width=\linewidth,
                height=5cm,
                keepaspectratio
            ]{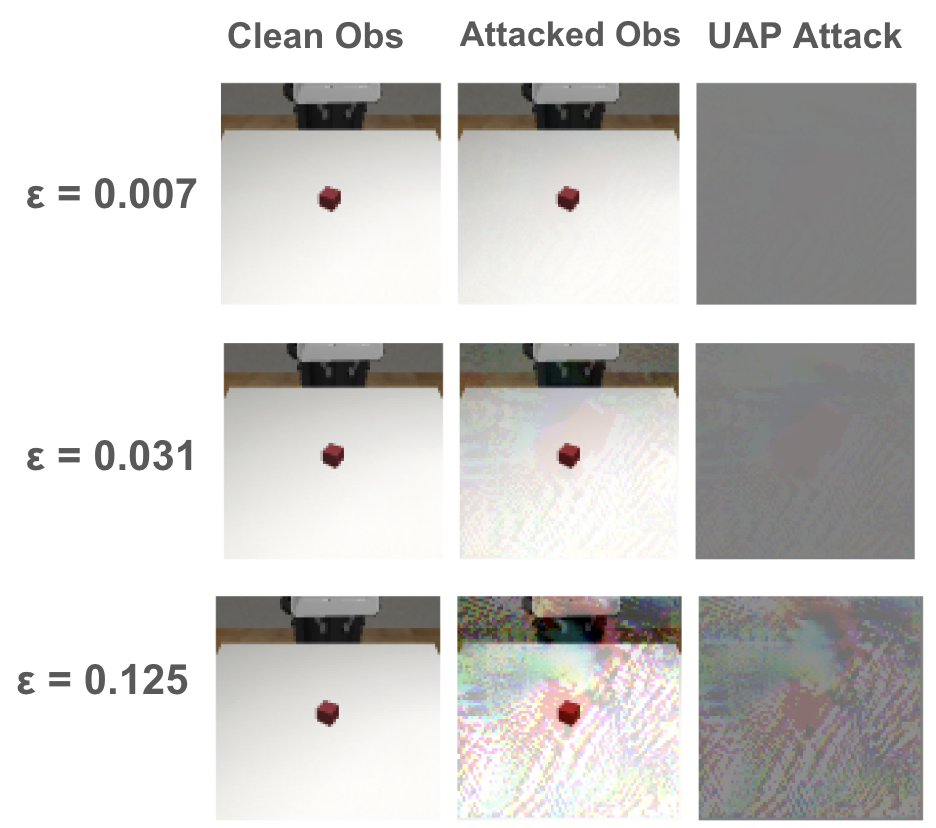}
        }
        \caption{Qualitative visualization of universal adversarial
        perturbations at increasing perturbation budgets. Columns show the
        clean agent-view observation, the corresponding perturbed observation,
           and the generated universal perturbation for
        $\epsilon \in \{0.007, 0.031, 0.125\}$. 
        Larger perturbation budgets
        produce increasingly visible changes in the policy's observation.}
        \label{fig:qualitative_eps}
    \end{subfigure}
    \hfill
    \begin{subfigure}[t]{0.48\textwidth}
        \vspace{0pt}
        \centering
        \parbox[c][5.2cm][c]{\linewidth}{%
            \centering
            \includegraphics[
                width=\linewidth,
                height=5cm,
                keepaspectratio
            ]{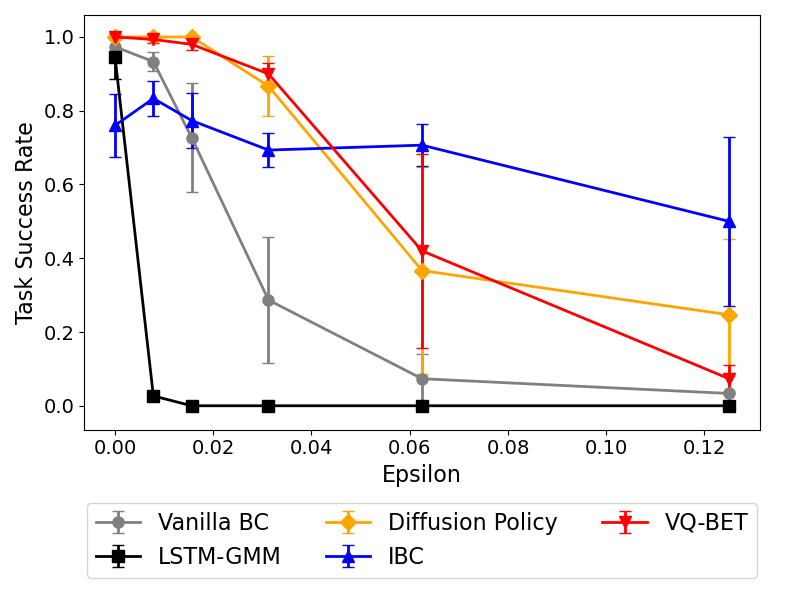}
        }
        \caption{Task success rates under increasing perturbation budgets
        ($\epsilon$) on the Lift task for different BC algorithms. Results are
        averaged over three seeds, with 50 rollouts per seed, and standard
        deviations are shown as error bars. Most algorithms exhibit sharp
        performance degradation even at small $\epsilon$, whereas IBC remains
        relatively more robust.}
        \label{fig:eps_sensitivity}
    \end{subfigure}

    \caption{Qualitative and quantitative evaluation of universal adversarial
    perturbations under different perturbation budgets.}
    \label{fig:eps_combined}
\end{figure*}


Our findings, as illustrated in Table~\ref{table:UAP_attack_results}, reveal significant vulnerabilities in the adversarial robustness of current algorithms when faced with perturbations in the observation space.

Classical explicit BC algorithms, such as Vanilla BC and LSTM-GMM exhibit high vulnerability to UAP attacks. 
This is evidenced by the task success rate in Lift (first column in Table \ref{table:UAP_attack_results}) dropping from unattacked rate of 0.97 to 0.07 under UAP attack for Vanilla BC and from 0.95 to 0 for LSTM-GMM.
We observe a similar trend across Can, and Square as well.
In contrast, algorithms employing iterative methods for action selection, such as IBC, exhibited relatively higher robustness in Lift. 
However, since IBC has lower unattacked task performance in Can and Square as compared to other algorithms in these tasks, it shows more robustness and exhibits almost similar performance under UAP attacks too. 
We conjecture that this enhanced resilience  of implicit algorithms can be attributed to the inherent stochasticity in their action selection processes during inference.

\begin{table*}[!t]
\centering
\caption{\textbf{Inter-Algorithm Black-Box and Within-Algorithm Grey-Box Transferability of UAP attacks on Lift, Can, and Square}: 
Each row shows perturbations crafted by the attacker algorithm, applied to target policies in columns under the same task. Each cell represents average task success rates of target policies under UAPs generated from
different attacker policies within the same task. Diagonal entries quantify within-algorithm vulnerability under grey box setting, while off-diagonal entries measure cross-algorithm transferability under black box setting. 
We also include random attacks as a baseline to assess whether the observed performance degradation is due to cross-algorithm transferability rather than random perturbations.
All transferability runs are evaluated on 10 rollouts for 6 seeds (total 60 runs) for diagonal grey box and for 9 seeds (total 90 runs) for each non-diagonal black box transfer.
To save space, we abbreviated the names of the different BC algorithms across the second header. For this table only, we abbreviate Vanilla BC to BC and LSTM to LSTM-GMM, while keeping IBC, DP, and VQ-BET consistent with the rest of the paper.
Lower values indicate more successful transfer attacks, as they correspond to lower target-policy success rates.
Diffusion Policy and VQ-BET achieve high task success rates under unattacked settings across all 3 tasks, but are quite vulnerable to black-box UAP attacks. 
}
\large
\renewcommand{\arraystretch}{1.5}
\resizebox{\textwidth}{!}{%
\begin{tabular}{|c|ccccc|ccccc|ccccc|}
\hline
& \multicolumn{5}{|c|}{\textbf{Lift}} 
& \multicolumn{5}{|c|}{\textbf{Can}} 
& \multicolumn{5}{|c|}{\textbf{Square}} \\
\diagbox{{\textbf{Attacker}}}{{\textbf{Target Policy}}} 
& \textbf{BC} & \textbf{LSTM} & \textbf{IBC} & \textbf{DP} & \textbf{VQ}
& \textbf{BC} & \textbf{LSTM} & \textbf{IBC} & \textbf{DP} & \textbf{VQ}
& \textbf{BC} & \textbf{LSTM} & \textbf{IBC} & \textbf{DP} & \textbf{VQ} \\
\noalign{\hrule height 1.2pt}
\textbf{Random} 
& 1.00(0.00) & 0.41(0.36) & 0.76(0.12) & 1.00(0.00) & 1.00(0.00)
& 0.72(0.17) & 0.83(0.09) & 0.02(0.04) & 0.99(0.03) & 0.93(0.09)
& 0.64(0.13) & 0.41(0.11) & 0.00(0.00) & 0.96(0.05) & 0.68(0.09) \\
\noalign{\hrule height 1.2pt}
\textbf{Vanilla BC} 
& 0.77(0.31) & 0.34(0.38) & 0.77(0.12) & 1.00(0.00) & 0.99(0.03)
& 0.02(0.04) & 0.48(0.32) & 0.02(0.04) & 0.31(0.24) & 0.91(0.07)
& 0.07(0.05) & 0.04(0.05) & 0.00(0.00) & 0.93(0.07) & 0.47(0.18) \\
\noalign{\hrule height 1.2pt}
\textbf{LSTM-GMM}
& 1.00(0.00) & 0.23(0.30) & 0.78(0.15) & 1.00(0.00) & 0.99(0.03)
& 0.24(0.23) & 0.48(0.21) & 0.03(0.05) & 0.61(0.23) & 0.83(0.12)
& 0.47(0.02) & 0.00(0.00) & 0.01(0.03) & 0.89(0.07) & 0.41(0.13) \\
\noalign{\hrule height 1.2pt}
\textbf{IBC}
& 1.00(0.00) & 0.11(0.19) & 0.83(0.09) & 0.99(0.03) & 1.00(0.00)
& 0.47(0.29) & 0.57(0.24) & 0.03(0.05) & 0.88(0.14) & 0.84(0.21)
& 0.63(0.11) & 0.01(0.03) & 0.00(0.00) & 0.90(0.08) & 0.41(0.13)\\
\noalign{\hrule height 1.2pt}
\textbf{DP}
& 0.93(0.08) & 0.16(0.21) & 0.77(0.13) & 0.97(0.05) & 1.00(0.00)
& 0.01(0.03) & 0.11(0.11) & 0.02(0.04) & 0.00(0.00) & 0.50(0.21)
& 0.04(0.05) & 0.00(0.00) & 0.01(0.03) & 0.12(0.19) & 0.22(0.12) \\
\noalign{\hrule height 1.2pt}
\textbf{VQ-BET}
& 0.96(0.07) & 0.19(0.26) & 0.79(0.14) & 1.00(0.00) & 1.00(0.00)
& 0.46(0.20) & 0.52(0.16) & 0.00(0.00) & 0.50(0.22) & 0.30(0.14)
& 0.43(0.15) & 0.01(0.03) & 0.00(0.00) & 0.87(0.11) & 0.10(0.18) \\
\hline
\end{tabular}%
}
\label{table:uap_transfer_algos}
\end{table*}

It is important to note that the effectiveness of these attacks varies depending on the complexity of the task environment. In more challenging environments such as \textit{Can} and \textit{Square}, universal adversarial perturbations reduce task success rates by over 90 \% across all algorithms, indicating severe vulnerability.
For instance, the Lift environment allows for a larger margin of error, making it more forgiving to substantial perturbations in actions. 
However, as task complexity increases, we observe a dramatic reduction in the robot task success rates (i.e., increase in attack success rates) across all algorithms. 
Mandlekar et al.~\cite{Mandlekar2021WhatMI} categorized the difficulty of the tasks with \textit{Lift} being the easiest, \textit{Can} being harder than \textit{Lift}, and \textit{Square} being harder than \textit{Can}. 

\subsection{\textbf{How sensitive are attacks to adversarial perturbation budget $\varepsilon$?}}\label{eps_sensitivity}
We systematically vary the perturbation budget $\varepsilon$ to study the impact of the attack budget on these modern BC algorithms.
We vary the perturbation budget by a factor of 2 ranging from 0 to 32/256 (0.125).

As shown in Figure \ref{fig:qualitative_eps}, the UAP is barely perceptible at low perturbation budgets and becomes increasingly visible as $\varepsilon$ increases.
Our analysis in Figure \ref{fig:eps_sensitivity} reveals that vulnerabilities in all behavior cloning algorithms exist even with minimal perturbations in our UAP attacks.
The task success rate is degraded quite drastically in classical explicit BC algorithms such as Vanilla BC and LSTM-GMM while the recent explicit BC algorithm, VQ-BET exhibits comparatively more robustness.
This heightened sensitivity to small perturbations highlights a concerning vulnerability in current behavior cloning approaches, suggesting that even well-constrained adversarial attacks can significantly compromise policy performance.

Interestingly, we find that implicit policies are significantly more robust to adversarial perturbations than explicit policies under white-box UAP attacks.

Among implicit methods, IBC displays the highest robustness, as it undergoes the smallest decrease in task success rate. 
We believe that this stems from IBC’s contrastive sampling, which provides redundancy against fixed perturbations. In contrast, Diffusion Policy employs a stochastic multi-step denoising process in which the fixed UAP perturbation is propagated through the entire denoising chain in a coherent way, leading to a more consistent shift in the resulting action sequence. 
These differences suggest that each model's inference dynamics play a central role in shaping its vulnerability.




\subsection{\textbf{Can adversarial examples transfer across different algorithms to enable black-box attacks?}}

The transferability of adversarial examples across different behavior cloning algorithms presents an intriguing phenomenon to study, especially given the substantial differences in their loss functions and training methodologies (as detailed in Section \ref{sec:methodology}). 
While these algorithms share a common image encoder (ResNet-18), their end-to-end training approaches may influence how distinct feature representations are used for action prediction. 
We conduct black-box transferability attacks to investigate whether the UAP attack crafted for one LfD algorithm generalizes to other algorithms within the same task.
Given a task and a known BC algorithm, we also study grey-box UAP attacks, where the attacker only knows about the choice of BC algorithm but does not have access to actual policy parameters and weights. 
We also compare against a random baseline, where perturbations are sampled uniformly from $[-\varepsilon,\varepsilon]$ for each camera view independently.

The results are shown in Table ~\ref{table:uap_transfer_algos}. In simpler environments like the Lift task, where baseline success rates are high ($>$90\% for most algorithms), we observed limited transferability with relatively small proportional drops in performance, aligning with our initial expectations. 
 
As we progressed to more complex environments such as Can and Square, in which the baseline success rates are themselves lower and the tasks are naturally less robust to action perturbations, we noticed that transferred attacks often caused larger proportional drops in performance relative to the baseline. 
For instance, Row 2 of Table \ref{table:uap_transfer_algos} shows that a UAP attack crafted using Vanilla BC reduces VQ-BET's task success rate to 0.99 in \textit{Lift}, to 0.91 in \textit{Can} and degrading even further to 0.47 in \textit{Square}.
Since IBC starts from a low performance baseline on Can and Square, grey-box and black-box transfer attacks cause only limited additional degradation while on Lift, its performance remains largely unchanged across the unattacked, black-box, and grey-box settings.
In line with this observation, out of all the algorithms we analyzed, DP and VQ-BET show a general trend of higher robustness to black-box transfer attacks even in complex tasks such as Can and Square.

Examining grey-box settings, the diagonal entries in Table \ref{table:uap_transfer_algos} indicate that grey-box transfer attacks are generally the strongest, causing the largest drops in performance. 
This aligns with our intuition that prior knowledge of the type of BC algorithm enables more effective attack construction even without access to actual model weights.

Compared to a random baseline, we also observe that black-box transfer attacks outperform random attacks in most cases, especially as the task complexity increases from Lift to Can to Square. Surprisingly, Vanilla BC along with DP and VQ-BET are most robust to random attacks across all tasks while LSTM-GMM shows the most vulnerability to random attacks. 

Overall our results underscore that even with little to no knowledge of the BC framework, effective attacks can be constructed, making transferability a critical concern for real-world deployment.





\section{CONCLUSION}
We present the first systematic comparison of the vulnerability of different Behavior Cloning (BC) algorithms to white-box, grey-box, and black-box adversarial attacks. 
Our experiments empirically show that pre-computed offline visual perturbations are sufficient to degrade the performance of five popular BC policies.
White-box attacks reduce average task success by at least 65\% on every evaluated task.
Interestingly, we find evidence that implicit policies such as IBC and Diffusion Policy and transformer-based policies such as VQ-BET are more robust than the explicit BC policies.
However, our results also indicate that the attack success rate is dependent on the task. As tasks get harder, it becomes easier to attack these algorithms.
We also demonstrate that Universal Adversarial Perturbation attacks exhibit cross-algorithm transferability, with larger relative performance drops in challenging environments. 
Among target policies with at least 50\% unattacked success, the strongest black-box transfers reduce task success by up to 88\% on Lift, 99\% on Can, and 100\% on Square.
We believe that our work lays the foundation for future work in the direction of adversarial robustness of BC policies. 
While much progress on adversarial robustness has been made in computer vision, the sequential nature of robotic BC policies and the complex relationship between vision representations and resulting actions (especially for implicit policies such as IBC and Diffusion Policy) provide an underexplored and exciting area for future research into both vulnerabilities and defenses. 
Future work should explore whether transferability arises from shared visual features, architectural similarity, loss-function similarity, or task-level behavioral similarities. 
Future work should also investigate different defense techniques for UAP attacks on BC policies such as adversarial training, input randomization, and encoder regularization.

\bibliographystyle{IEEEtran} 
\bibliography{refs} 

\end{document}